\pdfoutput=1

\documentclass[10pt,twocolumn,letterpaper]{article}

\usepackage{cvpr}              

\usepackage{cuted} 

\usepackage{currfile} 

\usepackage{caption} 

\usepackage{microtype}

\usepackage{placeins}

\usepackage{booktabs}
\usepackage{adjustbox}
\usepackage{multirow}
\usepackage{xcolor}
\definecolor{refgray}{gray}{0.33}
\usepackage{amsmath,amssymb}
\usepackage{xspace}

\definecolor{cvprblue}{rgb}{0.21,0.49,0.74}
\usepackage[pagebackref,breaklinks,colorlinks,allcolors=cvprblue]{hyperref}
\usepackage[capitalize]{cleveref}
\crefname{section}{Sec.}{Secs.}
\Crefname{section}{Section}{Sections}
\crefname{table}{Tab.}{Tabs.}
\Crefname{table}{Table}{Tables}

\def\confName{CVPR}
\def\confYear{2027}

\newcommand{\bubothmean}{56.8}

\newcommand{\buclinicalmean}{61.6}

\newcommand{\bufinalmean}{64.6}

\newcommand{\bufirstclinical}{6}
\newcommand{\bufirstfinal}{8}

\newcommand{\bufmgap}{+4.6}

\newcommand{\bufmmean}{60.0}

\newcommand{\bulearnedgainphysiomi}{+10.2}

\newcommand{\bulearnedmean}{63.9}

\newcommand{\buplaingap}{+8.8}

\newcommand{\buvmtdpmean}{55.8}

\newcommand{\buwavemean}{50.5}

\newcommand{\dispadftdcar}{40.2}

\newcommand{\dispadftdref}{25.0}

\newcommand{\dispfmoneadftdname}{CBraMod}
\newcommand{\dispfmoneadftdref}{2.2}
\newcommand{\dispfmoneadftdsel}{29.1}
\newcommand{\dispfmonetuslname}{REVE}
\newcommand{\dispfmonetuslref}{38.5}
\newcommand{\dispfmonetuslsel}{57.0}

\newcommand{\disptuslcar}{43.2}

\newcommand{\disptuslref}{48.4}

\newcommand{\disptusltcp}{71.1}
\newcommand{\disptuslwhite}{84.1}

\newcommand{\interimnote}{}

\newcommand{\linparamsmin}{3{,}074}

\newcommand{\mlpparamsmax}{1.2M}

\newcommand{\rcfcycworst}{33.7}
\newcommand{\rcfcycworsttask}{slowing}

\newcommand{\rcn}{8}

\newcommand{\rcrandgapbest}{36.6}
\newcommand{\rcrandgapbesttask}{slowing}

\newcommand{\rcrandgapmean}{15.2}

\newcommand{\rcrollBGmean}{-0.3}

\newcommand{\rcrolldown}{6}

\newcommand{\rcrollworst}{29.7}
\newcommand{\rcrollworsttask}{slowing}

\newcommand{\rcshufdown}{6}

\newcommand{\rcshufneg}{6.3}

\newcommand{\refcmadftd}{0.91}

\newcommand{\stdn}{8}

\newcommand{\stdrankours}{1.3}
\newcommand{\stdrankreve}{2.9}

\title{Your Vision Foundation Model Is Secretly an EEG Foundation Model}

\author{Guoan Wang \qquad Feng Liu\thanks{Corresponding author.}\\
Department of Industrial and Systems Engineering\\
School of Engineering, Rutgers University--New Brunswick\\
{\tt\small feng.liu@rutgers.edu}
}

\begin{document}
\maketitle
\begin{strip}
\centering
\includegraphics[width=0.99\linewidth]{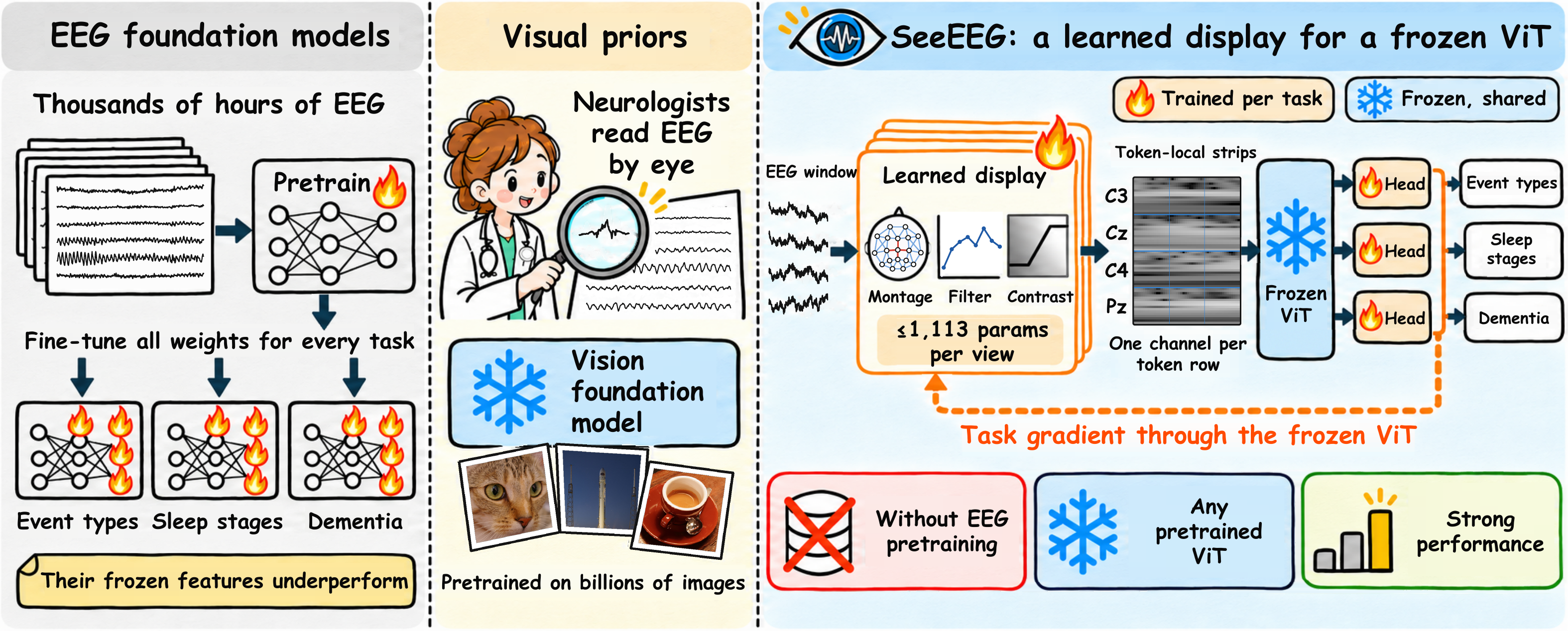}
\captionsetup{hypcap=false}
\captionof{figure}{\textbf{Learning the display through a frozen vision foundation model.} EEG foundation models, pretrained on up to tens of thousands of hours of recordings, typically require fine-tuning for each task because their frozen features fall short (left). Neurologists read EEG by visual inspection, and vision foundation models are pretrained on billions of images (middle). SeeEEG shares one frozen ViT backbone across tasks and trains only a lightweight display and readout head for each task (right).}
\captionsetup{hypcap=true}
\label{fig:teaser}%
\end{strip}

\begin{abstract}
Electroencephalography (EEG) foundation models are pretrained on thousands of hours of recordings, yet their frozen features transfer poorly to downstream tasks, and this transfer does not consistently improve with larger models or more pretraining data. We show that a vision foundation model that has never seen EEG is a stronger frozen EEG encoder than these specialized models, once the signals are rendered as images that a vision transformer can read. Our method, SeeEEG, keeps the vision backbone frozen and learns only how EEG is displayed to it. Each channel occupies a dedicated row of patches and is drawn either as a waveform at a fixed gain or as a scalogram whose frequency range spans the patch height, so that task-relevant patterns fall within individual tokens. The montage, filtering, and contrast parameters of this display are initialized from a standard clinical montage and optimized with the task loss through the frozen backbone, which amounts to at most 1,113 trainable parameters per view. We compare SeeEEG with seven EEG foundation models on eight clinical, sleep, affective, and brain-computer interface datasets, with every encoder frozen and paired with identical readout heads. With a frozen DINOv3 ViT-B, SeeEEG outperforms every EEG foundation model in balanced accuracy on all eight datasets and reaches a mean rank of \stdrankours\ in Cohen $\kappa$ or AUROC, compared with \stdrankreve\ for the strongest baseline, REVE. The learned displays transfer without further optimization to the video encoder V-JEPA 2.1, which outperforms every EEG foundation model on both metrics on six of the eight datasets. Replacing DINOv3 with a randomly initialized ViT-B lowers the score by \rcrandgapmean\ points on average, so the ability to read rendered EEG comes from visual pretraining.
\end{abstract}

\section{Introduction}
\label{sec:intro}

Electroencephalography (EEG) foundation
models~\cite{kostas2021bendr,yang2023biot,jiang2024labram,wang2024eegpt,wang2025cbramod,zhou2025csbrain,elouahidi2025reve}
are pretrained on large collections of unlabeled recordings, aiming to learn representations general enough
that downstream tasks require only a lightweight classifier atop the frozen encoder to achieve strong performance. Acquiring such representations demands substantial data, and most models in \cref{tab:pretrain} are pretrained on thousands
of hours of EEG recordings. Even at this scale, recent benchmarks show that their frozen features transfer poorly to many downstream
tasks~\cite{xiong2026eegfmbench,wu2025adabrain,liu2026compass,yang2026worth} and do not reliably improve with increased model capacity or additional
pretraining data~\cite{xiong2026eegfmbench,liu2026compass,yang2026worth,lee2025brainwave}. These findings suggest that learning broadly transferable frozen EEG representations remains an open challenge.

\begin{table}[t]
\centering
\footnotesize
\setlength{\tabcolsep}{4pt}
\caption{\textbf{Pretraining data of EEG foundation models.} Entries are as reported in the original papers.}
\label{tab:pretrain}
\begin{adjustbox}{max width=\linewidth}
\begin{tabular*}{\linewidth}{@{\extracolsep{\fill}}llr}
\toprule
Model & Pretraining corpus & Size \\
\midrule
BENDR~\cite{kostas2021bendr} & TUEG & 1.5\,TB \\
BIOT~\cite{yang2023biot} & SHHS + PREST & 10.2M windows \\
LaBraM~\cite{jiang2024labram} & about 20 datasets & 2,535\,h \\
EEGPT~\cite{wang2024eegpt} & 5 datasets & 279 subjects \\
CBraMod~\cite{wang2025cbramod} & TUEG & $>$9,000\,h \\
CSBrain~\cite{zhou2025csbrain} & TUEG & $>$9,000\,h \\
REVE~\cite{elouahidi2025reve} & 92 datasets & 60,000\,h \\
\bottomrule
\end{tabular*}
\end{adjustbox}
\end{table}

Vision foundation models are trained at an even larger scale. DINOv3~\cite{simeoni2025dinov3}, for example, is
pretrained on 1.7 billion images, and its frozen features outperform specialized
models across a broad range of vision tasks. Models of this kind encode a strong prior for recognizing visual
patterns. This prior is directly relevant to EEG analysis, because neurologists interpret EEG recordings primarily through visual inspection, recognizing spikes, periodic
discharges, spindles, and artifacts by their waveforms on a page of traces. Prior work has applied frozen vision models to time series rendered as images for classification~\cite{roschmann2025tivit} and
forecasting~\cite{chen2024visionts}. This cross-modal transfer aligns with the Platonic representation
hypothesis~\cite{huh2024platonic}, which posits that models trained across different modalities converge toward increasingly similar
representations as they scale, because images, text, and other modalities are all observations of the same underlying world.
Recent theoretical work proves this convergence for joint-embedding predictive architectures under idealized
conditions~\cite{balestriero2026platonic}. These observations motivate our central question: \emph{can a vision foundation model, pretrained on far more data than EEG
foundation models and already proven across a wide range of visual tasks, serve as a general encoder for EEG with a learnable input module that adapts EEG signals to the frozen model?}

Existing image-based approaches for EEG either still require training on EEG data or process each channel in isolation.
MTDP~\cite{li2026mtdp} probes frozen DINOv3
features extracted from the raw channel-by-time signal matrix rendered as a grayscale image, yet its final encoder is distilled from DINOv3
and pretrained on EEG recordings. BraVista, EEGiT,
and VIPEEGNet fine-tune their vision encoders on EEG images~\cite{azizeddin2026bravista,zhou2026eegit,sun2025vipeegnet},
whereas TiViT~\cite{roschmann2025tivit} processes each channel as a separate image, preventing spatial relationships across electrodes from
appearing within a single image. In contrast, we keep the vision foundation model completely frozen, without any EEG pretraining, distillation, or
fine-tuning. We enable it to interpret EEG signals only by learning how EEG is displayed to the backbone, together with a lightweight task head.

We term this approach SeeEEG and outline it in \cref{fig:teaser,fig:method}. Its design follows how a ViT processes an
image: the local content of each patch is embedded directly, whereas relationships between patches are captured
only through attention tuned to natural images. SeeEEG therefore renders each channel in a dedicated strip of
patches, either as grayscale amplitude at a fixed gain or as a scalogram whose frequency range spans the patch
height, so that task-relevant patterns fall within individual tokens. The content of these strips is set by
a display module parameterized by the montage, filtering, and contrast settings of a clinical EEG viewer, in which a bipolar or Laplacian montage
places the potential difference between neighboring electrodes within a single token. Initialized from
a standard clinical montage, the display module is optimized with the task loss through the frozen backbone, using at most 1,113
trainable parameters per view. We
evaluate SeeEEG against seven EEG foundation models on \stdn\ datasets, with all encoders frozen and paired with identical readout heads
on the same samples. Our main contributions are summarized as follows:
\begin{itemize}
\item \textbf{Learning the display through a frozen vision foundation model.} We treat the montage, filtering, and contrast
settings of a clinical EEG viewer as learnable inputs of a frozen ViT and optimize them with the task loss, so the learned
display remains a valid page of EEG traces.
\item \textbf{Token-local rendering.} We render each channel in a dedicated row of tokens, so every token holds one channel
over 16 samples, either as amplitude at a fixed gain or as its spectrum from 1 to 64\,Hz.
\item \textbf{Vision pretraining in place of EEG pretraining.} Across \stdn\ datasets, a frozen DINOv3 ViT-B outperforms every
EEG foundation model in balanced accuracy and achieves a mean rank of \stdrankours, compared to \stdrankreve\ for the
strongest baseline, REVE. The learned displays transfer to V-JEPA 2.1 and a supervised ViT.
\end{itemize}
\section{Related work}
\label{sec:related}

\paragraph{EEG foundation models.}
EEG foundation models are transformers pretrained on large collections of unlabeled EEG recordings, differing
primarily in their pretraining objectives and in how they tokenize channels and time. BENDR learns contrastive
representations of raw EEG~\cite{kostas2021bendr}, BIOT tokenizes each channel independently~\cite{yang2023biot}, LaBraM
predicts the codes of a learned neural tokenizer for masked patches~\cite{jiang2024labram}, EEGPT combines masked
reconstruction with spatio-temporal alignment~\cite{wang2024eegpt}, CBraMod applies criss-cross attention to masked
patches~\cite{wang2025cbramod}, CSBrain models EEG across temporal and spatial scales~\cite{zhou2025csbrain}, and REVE
pretrains on arbitrary electrode layouts using 4D positional encodings~\cite{elouahidi2025reve}. All seven models learn their
representations exclusively from EEG data. Benchmarks evaluating these models under unified protocols show that the frozen
features of most models transfer poorly to many downstream
tasks~\cite{xiong2026eegfmbench,wu2025adabrain,liu2026compass,yang2026worth} and cannot yet serve as general-purpose fixed feature extractors~\cite{liu2026compass,yang2026worth}.
Neither larger model capacities nor additional pretraining data reliably improve downstream
performance~\cite{xiong2026eegfmbench,liu2026compass,yang2026worth,lee2025brainwave}.

\paragraph{Vision foundation models for signals and EEG.}
The frozen features of vision foundation models transfer effectively to downstream tasks via a linear
probe~\cite{radford2021clip,oquab2024dinov2,simeoni2025dinov3} and also generalize to non-visual signals rendered as images.
AST and ViTST fine-tune pretrained vision transformers on audio spectrograms and line graphs of irregularly sampled
time series, respectively~\cite{gong2021ast,li2023vitst}. VisionTS performs zero-shot forecasting using an image-pretrained masked autoencoder that
processes time series rendered as images~\cite{chen2024visionts}, whereas TiViT feeds each channel into a frozen ViT as a separate
image~\cite{roschmann2025tivit}. In the context of EEG, MTDP probes frozen DINOv3 features extracted from the signal matrix and distills them
into an encoder pretrained on EEG~\cite{li2026mtdp}, while BraVista, EEGiT, and VIPEEGNet fine-tune vision encoders directly on EEG
images~\cite{azizeddin2026bravista,zhou2026eegit,sun2025vipeegnet}. Across all these approaches, the encoder is either trained on the target signal modality or
restricted to processing each channel as a separate image. SeeEEG keeps the vision foundation model completely frozen, without any EEG
pretraining, distillation, or fine-tuning. All channels are rendered within a single image, one channel per row of tokens, and SeeEEG
learns only how EEG signals are displayed to the backbone through the montage, filtering, and contrast settings of a clinical EEG viewer, alongside a
lightweight task head.

\section{Method}
\label{sec:method}

\begin{figure*}[t]
\centering
\includegraphics[width=0.99\linewidth]{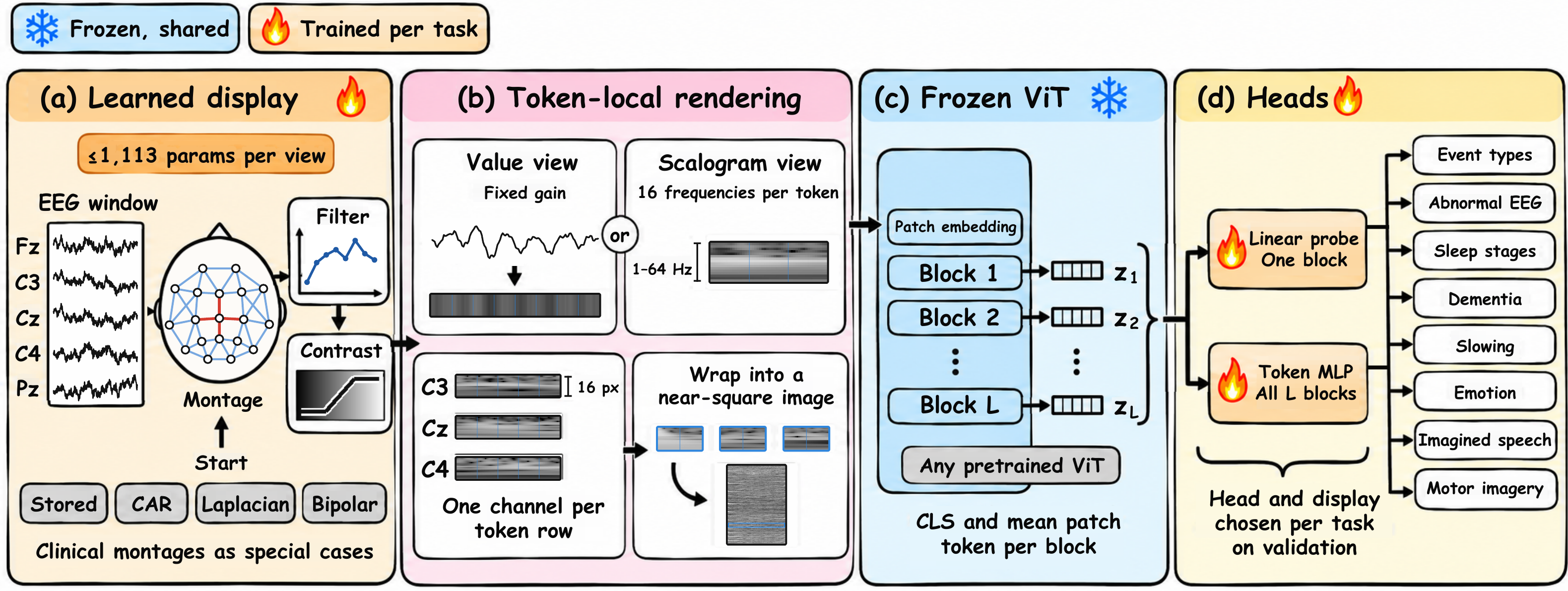}
\caption{\textbf{Overview of SeeEEG.} (a) The learned display sets the montage, display filter, and contrast, starting from one of the clinical montages that it contains as special cases. (b) Each channel fills a dedicated row of 16-pixel tokens in the value or scalogram view, and long windows are wrapped into a near-square image. (c) A frozen pretrained ViT extracts the class token and the mean patch token of each transformer block. (d) A linear probe or a token MLP reads these tokens, and the readout head and display are selected on validation for each task.}
\label{fig:method}
\end{figure*}

SeeEEG turns a frozen vision foundation model into an EEG encoder by learning how EEG signals are displayed to the backbone, as illustrated in
\cref{fig:method}. Each input sample is a recording window $x\in\mathbb{R}^{C\times T}$ with $C$ channels and $T$ time steps at 200\,Hz. A
linear transformation $D_\omega$ derives and filters the displayed channels, a rendering operator $R_\phi$ converts them into an
image, a frozen ViT $f$ encodes the image, and a lightweight readout head $h$ predicts the label:
\begin{equation}
\hat y = h\big(f(R_\phi(D_\omega x))\big).
\label{eq:pipeline}
\end{equation}
Only the display parameters $\theta=(\omega,\phi)$ and the head $h$ are trained, and gradients pass through the frozen backbone $f$ only to update $\theta$.

\subsection{Token-local rendering}
\label{sec:render}

The backbone $f$ is a ViT pretrained on natural images. It partitions an input image $I\in\mathbb{R}^{3\times H\times W}$ into a
grid of $P\times P$ patches $I_{mn}$ and projects each patch with a shared linear mapping before $L$ transformer blocks:
\begin{equation}
e_{mn} = E\,\mathrm{vec}(I_{mn}), \qquad E\in\mathbb{R}^{D\times 3P^2},
\label{eq:patch}
\end{equation}
where $D$ is the token dimension. With rotary or interpolated learned position embeddings, $f$ accepts any image size divisible by $P$.
A pattern within one patch enters the representation directly through $E$, whereas dependencies between patches are captured only by
self-attention tuned to the layout of objects and parts in natural images. We therefore render EEG so that task-relevant structures
lie within individual tokens. Each displayed channel $\tilde x_c$ of
$\tilde x=D_\omega x\in\mathbb{R}^{\tilde C\times T}$ is rendered as a horizontal strip of $P$ pixel rows aligned with the patch grid,
with one time sample per pixel column, so that every token holds one channel over $P$ consecutive samples. The
channel-by-time signal matrix~\cite{li2026mtdp} instead packs $P$ channels into each token row, so every token mixes several electrodes.

Each strip shows its channel in one of two views. The value view draws each sample as a grayscale intensity repeated over
the $P$ rows of the strip:
\begin{equation}
v_{c}(t) = \mathrm{clip}\Big(\tfrac12+\frac{\gamma\,\tilde x_c(t)}{2A}\Big),
\label{eq:value}
\end{equation}
where $\mathrm{clip}$ restricts values to $[0,1]$, $A$ is a fixed amplitude scale estimated from the training windows of
each dataset, and $\gamma$ is a learnable gain. Unlike per-window min-max normalization, the fixed
gain preserves the absolute amplitude on which clinical interpretation relies.

The scalogram view fills row $j=1,\dots,P$ with the power of the channel at frequency $\nu_j$:
\begin{equation}
u_{cj}(t) = \mathrm{clip}\Big(\tfrac12+\frac{a_j\,z_{cj}(t)+b_j}{6}\Big),
\label{eq:scalogram}
\end{equation}
where $z_{cj}$ is the standardized log-power of the Morlet wavelet transform~\cite{torrence1998wavelet} of $\tilde x_c$ at $\nu_j$
and $(a_j,b_j)$ are learnable contrast parameters. The $P$ frequencies are logarithmically spaced from 1 to
64\,Hz over the full patch height, so every token holds the spectrum of one channel over $P$ samples. A long window is split into $k^\star=\arg\min_k\big|\log\big(P\tilde Ck/\lceil T/k\rceil\big)\big|$ consecutive
segments, stacked vertically as further strips of the same channel to keep the image close to square.
The grayscale image is copied to the three color channels and normalized with ImageNet statistics.

\subsection{From clinical montages to learned displays}
\label{sec:learn}

For a frozen ViT, a bipolar or Laplacian derivation places the potential difference between neighboring electrodes
within a single strip and thus within individual tokens. The stored referential channels, the common average reference (CAR),
the surface Laplacian~\cite{hjorth1975laplacian}, and the bipolar double-banana montage used to annotate TUH
recordings~\cite{lopez2016montages,golmohammadi2019tuev} all belong to one family with a global reference term and local spatial
weights:
\begin{equation}
\tilde x_c = x_c - g_c\,\bar x - \sum_{j\in\mathcal{N}(c)} w_{cj}\,x_j ,
\label{eq:display}
\end{equation}
where $\bar x$ is the channel mean and $\mathcal{N}(c)$ holds the four electrodes nearest to $c$ on the 10-05
layout~\cite{oostenveld2001five}. The stored montage is $g=0,w=0$, the CAR sets $g_c=1,w=0$, the Laplacian sets $w_{cj}=1/4$, and a bipolar derivation $c-j$ adds $j$ to $\mathcal{N}(c)$ with $w_{cj}=1$.

Because the ViT does not know which electrode a strip shows, a strip can also show broader channel combinations
outside this family. We fit two such displays in closed form on the training windows of the clinical montage
selected on validation. The spatial-pattern display renders the one-versus-rest
common spatial patterns of the classes~\cite{ramoser2000csp,blankertz2008csp}, and
the whitened display applies ZCA whitening~\cite{kessy2018whitening} across channels, which decorrelates them
while keeping each strip dominated by its own channel. EEG foundation models that rely on electrode identity embeddings cannot read
such displays.

The learned display starts from the clinical montage selected on validation and optimizes the settings of a clinical EEG viewer.
The montage weights $(g,w)$ set the spatial contrast. A zero-phase display filter scales each frequency $\nu$ of
$\tilde x$ by a gain $G(\nu)$, whose logarithm is bounded by $\pm3$ and interpolated linearly in $\log\nu$ between eight knots spaced
one octave apart from 0.5 to 64\,Hz, and both views are rendered from the filtered signal. The contrast parameters
$(\gamma,\{a_j,b_j\})$ set the grayscale mapping. Optional row filters pass each of the $P$ rows of
a value strip through its own 33-tap filter, so that one token shows its channel through $P$ different
filters. Each scalogram row may also shift its center frequency and wavelet width by up to a factor of two, and
$\mathcal{N}(c)$ grows to the eight nearest electrodes. An optional color map turns the intensity of each strip row into a
color through a monotonic luminance curve and two chroma curves. The montage and filters form $\omega$, and the
contrast and color map form $\phi$. Because the montage and filters are linear in the signal, a learned
display is still a valid page of EEG traces rather than the output of a neural network prepended to the backbone.

Validation selects among four variants of the learned display. The first optimizes the montage, contrast, and display
filter, and the second and third add the color map and the row filters, respectively. These three start at the clinical display,
with unit filter gains and identity row filters, so the initial image is unchanged. The fourth variant
mixes the clinical and spatial-pattern montages $B_m$ as $\sum_m\mathrm{softmax}(\alpha)_m B_m$ and blends each value
row with the band-limited waveform at its scalogram frequency, initialized at their mean and an equal blend. The resulting
displays have 29 to 1,113 trainable parameters per view. Neurologists also adjust the display sensitivity for each recording,
which a parameter-free automatic sensitivity emulates by scaling the displayed signals of each window by
$m/\mathrm{median}\,|\tilde x|$, where $m$ is the median absolute amplitude of the training windows used to calibrate $A$. Because this discards the absolute amplitude kept by the fixed gain, validation adopts it for a
display only if it improves both the score of the selected readout head and the mean score over all candidate heads.

\subsection{Readout and display fitting}
\label{sec:fit}

After each block $b$ of $f$, we apply the final layer normalization and concatenate the class token with the mean patch token,
$z_b=[\,\mathrm{cls}_b;\ \overline{\mathrm{patch}}_b\,]$. We use two readout heads, a linear probe on a single
block and a token MLP that projects each of the $L$ block vectors to 128 dimensions and classifies their concatenation
through one hidden layer. Each head reads the value view, the scalogram view, or both, which gives six candidate heads per display.
Validation selects the block, regularization strength, readout head, and display variant by Cohen $\kappa$ for multi-class
tasks and AUROC for binary tasks. An encoder evaluated on several displays predicts with either its best display or
the mean class probabilities of the top three validation displays $\mathcal{S}_3$,
\begin{equation}
p(y\mid x)=\frac{1}{3}\sum_{d\in\mathcal{S}_3} p_d(y\mid x),
\label{eq:ensemble}
\end{equation}
whichever scores higher on validation. Every EEG foundation model evaluated on several montages follows the same protocol,
and no design choice uses test scores.

Optimizing the display through these readout heads invites two shortcuts. A display trained jointly with a head on the same
windows can fit subject-specific traits, because when the label is a subject-level property,
the signature of each subject predicts it. A display optimized through a fixed head can instead adapt to that
head, for instance by rescaling the feature dimensions the head weights most. To prevent both shortcuts, we cross-fit the display against
co-evolving readout heads. Each view is optimized separately through the readout architecture selected on validation for the initial
display, restricted to that view. This architecture is fitted $K=5$ times, each time without the training subjects of one
fold $\mathcal{F}_k$, and each training window updates the display only through the head that has not seen its
subject:
\begin{equation}
\min_{\theta}\;\sum_{k=1}^{K}\;\sum_{(x,y)\in\mathcal{F}_k}\ell\big(h_{-k}(f(R_\phi(D_\omega x))),\,y\big),
\label{eq:learn}
\end{equation}
where $\ell$ is the cross-entropy loss and $h_{-k}$ is the head fitted without fold $k$.

As the display evolves, each $h_{-k}$ keeps training on the data outside its fold, and linear heads standardize their inputs
with running statistics so that rescaling features gives no advantage. The display is thus rewarded only for patterns that a
head trained on other subjects can use, as required at test time. On within-subject benchmarks, which test held-out trials of the training subjects,
the folds split trials instead of subjects. After every epoch, the $K$ heads are
refitted on an exponential moving average of the display parameters, their mean prediction scores this
averaged display on validation, and the best epoch, including the unoptimized initial display, is kept. The learning
rate of each view and the display variant are chosen by the same criterion, with ties going to smaller rates and
simpler variants. The selected learned display is then frozen and enters the final display selection as one more candidate
beside the clinical and closed-form displays.

\section{Experimental setup}
\label{sec:setup}

\begin{table*}[t]
\centering
\footnotesize
\setlength{\tabcolsep}{3.2pt}
\caption{\textbf{Comparison of frozen encoders.} Test scores (\%, mean over three seeds). Every encoder is frozen and read out by the same two heads, with the head and display selected on validation. The SeeEEG rows differ only in the ViT-B/16 backbone and share the same displays. Superscripts denote the selected display (c: CAR, l: Laplacian, b: bipolar, w: whitened, m: learned, e: ensemble of the three best displays, s: automatic sensitivity), and no superscript denotes the stored or native montage. Bold and underline mark the best and second-best results. \interimnote}
\label{tab:main}
\begin{adjustbox}{max width=\linewidth}
\begin{tabular}{lcccccccccccccccc}
\toprule
 & \multicolumn{2}{c}{TUEV} & \multicolumn{2}{c}{TUAB} & \multicolumn{2}{c}{HMC} & \multicolumn{2}{c}{ADFTD} & \multicolumn{2}{c}{TUSL} & \multicolumn{2}{c}{SEED} & \multicolumn{2}{c}{BCIC2020-3} & \multicolumn{2}{c}{PhysioMI} \\
\cmidrule(lr){2-3}\cmidrule(lr){4-5}\cmidrule(lr){6-7}\cmidrule(lr){8-9}\cmidrule(lr){10-11}\cmidrule(lr){12-13}\cmidrule(lr){14-15}\cmidrule(lr){16-17}
 & B-Acc & $\kappa$ & B-Acc & AUROC & B-Acc & $\kappa$ & B-Acc & $\kappa$ & B-Acc & $\kappa$ & B-Acc & $\kappa$ & B-Acc & $\kappa$ & B-Acc & $\kappa$ \\
\midrule
\multicolumn{17}{c}{Frozen EEG foundation models} \\
\midrule
BENDR~\cite{kostas2021bendr} & 22.0 & 14.5 & 67.3 & 73.8 & 45.0 & 34.4 & 39.4 & 9.9 & 40.1 & 10.0 & 33.8 & 0.7 & 19.9 & -0.2 & 27.8 & 3.7 \\
BIOT~\cite{yang2023biot} & 43.2 & 59.3 & 75.9 & 83.5 & 64.5 & 58.5 & 40.8 & 15.5 & 53.1 & 21.8 & 51.4 & 27.6 & 21.6 & 2.0 & 25.9 & 1.2 \\
LaBraM~\cite{jiang2024labram} & 46.8$^{\mathrm{e}}$ & 55.9$^{\mathrm{e}}$ & 76.3 & 85.4 & 65.7 & 59.5 & 44.4$^{\mathrm{c}}$ & 21.9$^{\mathrm{c}}$ & 55.8$^{\mathrm{c}}$ & 28.7$^{\mathrm{c}}$ & 50.2$^{\mathrm{l}}$ & 25.8$^{\mathrm{l}}$ & 24.8$^{\mathrm{e}}$ & 5.9$^{\mathrm{e}}$ & 27.8 & 3.7 \\
EEGPT~\cite{wang2024eegpt} & 68.9$^{\mathrm{c}}$ & \underline{88.4}$^{\mathrm{c}}$ & 79.7 & 88.7 & 68.5 & 64.0 & 40.6 & 14.1 & 71.5 & 51.4 & 55.7$^{\mathrm{e}}$ & 34.1$^{\mathrm{e}}$ & 29.0$^{\mathrm{e}}$ & 11.3$^{\mathrm{e}}$ & 38.0 & 17.3 \\
CBraMod~\cite{wang2025cbramod} & 53.9$^{\mathrm{e}}$ & 60.2$^{\mathrm{e}}$ & 80.7$^{\mathrm{e}}$ & 88.9$^{\mathrm{e}}$ & 68.3$^{\mathrm{e}}$ & 64.0$^{\mathrm{e}}$ & 48.9$^{\mathrm{e}}$ & 29.1$^{\mathrm{e}}$ & 63.8$^{\mathrm{w}}$ & 42.7$^{\mathrm{w}}$ & 55.9$^{\mathrm{e}}$ & 34.4$^{\mathrm{e}}$ & 26.8$^{\mathrm{e}}$ & 8.5$^{\mathrm{e}}$ & 31.5$^{\mathrm{e}}$ & 8.7$^{\mathrm{e}}$ \\
CSBrain~\cite{zhou2025csbrain} & 53.1 & 56.8 & 77.0 & 85.8 & 64.7 & 59.8 & 36.9 & 7.9 & 71.4 & 52.2 & 49.5 & 24.6 & 27.0 & 8.7 & 30.8 & 7.8 \\
REVE~\cite{elouahidi2025reve} & 68.3$^{\mathrm{c}}$ & \textbf{88.9}$^{\mathrm{c}}$ & 79.7 & 88.2 & 68.4 & 63.1 & 51.9$^{\mathrm{l}}$ & 29.8$^{\mathrm{l}}$ & 74.5$^{\mathrm{c}}$ & 57.0$^{\mathrm{c}}$ & 61.5$^{\mathrm{e}}$ & 42.7$^{\mathrm{e}}$ & 23.6$^{\mathrm{c}}$ & 4.4$^{\mathrm{c}}$ & \underline{45.1} & \underline{26.8} \\
\midrule
\multicolumn{17}{c}{Frozen vision foundation models without EEG pretraining} \\
\midrule
SeeEEG, V-JEPA 2.1~\cite{murlabadia2026vjepa21} & \textbf{70.8}$^{\mathrm{m}}$ & 86.2$^{\mathrm{m}}$ & \underline{80.9}$^{\mathrm{e}}$ & \underline{89.3}$^{\mathrm{e}}$ & 71.7$^{\mathrm{e}}$ & 67.5$^{\mathrm{e}}$ & \underline{58.4}$^{\mathrm{c}}$ & \textbf{40.1}$^{\mathrm{c}}$ & \underline{76.7}$^{\mathrm{b}}$ & \underline{61.2}$^{\mathrm{b}}$ & \underline{64.5}$^{\mathrm{e}}$ & \underline{47.3}$^{\mathrm{e}}$ & \underline{33.1}$^{\mathrm{e}}$ & \underline{16.4}$^{\mathrm{e}}$ & 39.3$^{\mathrm{m}}$ & 19.1$^{\mathrm{m}}$ \\
SeeEEG, supervised IN-21k~\cite{steiner2022augreg} & 65.4$^{\mathrm{e}}$ & 85.3$^{\mathrm{e}}$ & 80.4$^{\mathrm{e}}$ & \textbf{89.5}$^{\mathrm{e}}$ & \underline{71.9}$^{\mathrm{e}}$ & \underline{67.7}$^{\mathrm{e}}$ & \textbf{59.1}$^{\mathrm{e}}$ & \underline{39.5}$^{\mathrm{e}}$ & 66.6$^{\mathrm{e}}$ & 44.5$^{\mathrm{e}}$ & 63.7$^{\mathrm{e}}$ & 46.1$^{\mathrm{e}}$ & 32.8$^{\mathrm{e}}$ & 16.1$^{\mathrm{e}}$ & 36.0$^{\mathrm{m}}$ & 14.7$^{\mathrm{m}}$ \\
SeeEEG, DINOv3~\cite{simeoni2025dinov3} & \underline{70.6}$^{\mathrm{e}}$ & 85.1$^{\mathrm{e}}$ & \textbf{81.2}$^{\mathrm{e}}$ & \textbf{89.5}$^{\mathrm{e}}$ & \textbf{72.0}$^{\mathrm{e}}$ & \textbf{67.8}$^{\mathrm{e}}$ & 54.1$^{\mathrm{c,s}}$ & 34.0$^{\mathrm{c,s}}$ & \textbf{90.4}$^{\mathrm{w}}$ & \textbf{84.1}$^{\mathrm{w}}$ & \textbf{65.6}$^{\mathrm{e}}$ & \textbf{49.0}$^{\mathrm{e}}$ & \textbf{35.6}$^{\mathrm{e}}$ & \textbf{19.4}$^{\mathrm{e}}$ & \textbf{47.2}$^{\mathrm{m,s}}$ & \textbf{29.7}$^{\mathrm{m,s}}$ \\
\bottomrule
\end{tabular}
\end{adjustbox}
\end{table*}

\paragraph{Datasets.}
Our evaluation spans eight datasets covering event-type classification on TUEV~\cite{harati2015tuev}, abnormal-EEG
detection on TUAB~\cite{lopez2015tuab,lopezdediego2017tuab,obeid2016tuh}, sleep staging on
HMC~\cite{alvarez2021hmc,alvarez2022hmcdata,goldberger2000physionet}, dementia classification on
ADFTD~\cite{miltiadous2023adftd}, slowing classification on TUSL~\cite{vonweltin2017tusl}, emotion recognition on
SEED~\cite{zheng2015seed}, imagined speech recognition on BCIC2020-3~\cite{jeong2022bcic2020}, and motor imagery classification on
PhysioMI~\cite{schalk2004bci2000,schalk2009eegmmidb,goldberger2000physionet}. For BCIC2020-3, we retain the official competition split. All remaining
datasets follow the preprocessing pipeline, windowing configurations, and subject splits established by Xiong \etal~\cite{xiong2026eegfmbench}, except that
we repartition the validation and test subjects for HMC. All windows are sampled at 200\,Hz.

\paragraph{Baselines.}
We compare SeeEEG against seven EEG foundation models: BENDR~\cite{kostas2021bendr}, BIOT~\cite{yang2023biot},
LaBraM~\cite{jiang2024labram}, EEGPT~\cite{wang2024eegpt}, CBraMod~\cite{wang2025cbramod}, CSBrain~\cite{zhou2025csbrain},
and REVE~\cite{elouahidi2025reve}. Each model is loaded with its released weights through the input adapters of
Xiong \etal~\cite{xiong2026eegfmbench} and kept frozen. After each transformer block, output tokens are mean- and
max-pooled and passed to our two readout heads with the same hyperparameter grids, random seeds, and validation criteria as SeeEEG. Validation also selects the
montage of each baseline among those compatible with its architecture. REVE, EEGPT, and LaBraM are evaluated on the stored referential, common-average, and Laplacian
montages. CBraMod, which does not embed channel identities, also supports the bipolar montage and the spatial-pattern and
whitened displays. BENDR and BIOT use their pretraining montages, and CSBrain uses the stored referential montage.

\paragraph{Backbones and evaluation.}
SeeEEG learns its displays through the DINOv3~\cite{simeoni2025dinov3} ViT-B/16 from the timm library~\cite{wightman2019timm}.
With $P=16$ at 200\,Hz, each token spans 80\,ms of one channel. To test whether the results depend on DINOv3, we replace it
with two other frozen ViT-B/16 backbones, the self-supervised video encoder V-JEPA 2.1~\cite{murlabadia2026vjepa21} and a
supervised ViT trained on ImageNet-21k~\cite{steiner2022augreg}, which read the same candidate displays, including the one
learned through DINOv3. We report balanced accuracy together with Cohen $\kappa$ for multi-class tasks and AUROC for binary
tasks, averaged over three seeds on the test split. The supplementary material details the statistics and preprocessing of each dataset, the input adapters of the baselines,
and the hyperparameters of the readout heads and display fitting.

\section{Results}
\label{sec:results}

\subsection{Comparison with EEG foundation models}
\label{sec:main}

All encoders in \cref{tab:main} are frozen and share the same readout heads, hyperparameter grids, and validation criteria. With a DINOv3 backbone,
SeeEEG outperforms every EEG foundation model in balanced accuracy on all eight datasets.
Ranked by $\kappa$ or AUROC, SeeEEG reaches a mean rank of \stdrankours, compared to
\stdrankreve\ for the strongest baseline, REVE. The margins are largest on tasks that rely on
spatial or spectral contrast, reaching $+27.1$ in $\kappa$ over the best EEG foundation model on slowing classification and
$+8.1$ on imagined speech recognition. The backbone has never seen EEG, whereas six of the seven
EEG foundation models saw unlabeled test recordings of some of these datasets during pretraining. SeeEEG still outperforms BENDR, CBraMod, and
CSBrain in both metrics on TUEV, TUAB, and TUSL, all three of which are contained in their TUH pretraining corpus.

This advantage is not specific to DINOv3. On the same displays, the self-supervised video encoder V-JEPA 2.1 outperforms
every EEG foundation model on both metrics on six of the eight datasets, and a supervised ImageNet-21k ViT does so
on four, although neither backbone was used to optimize the displays. In every configuration, the backbone stays fixed, and SeeEEG trains only a display module of
at most 1,113 parameters per view and a readout head of \linparamsmin\ to \mlpparamsmax\ parameters.

\subsection{Display ablation}
\label{sec:display}

\begin{table}[t]
\centering
\footnotesize
\setlength{\tabcolsep}{3.0pt}
\caption{\textbf{From plain images to the learned display.} Test balanced accuracy (\%, mean over three seeds) of the frozen DINOv3 ViT-B when only the input image varies. Plain images render each channel as a trace or as one pixel row of a matrix. The strips use the stored montage, the clinical montage selected on validation, or the learned display. Bold marks the best image for each dataset. \interimnote}
\label{tab:buildup}
\begin{adjustbox}{max width=\linewidth}
\begin{tabular}{lccccccc}
\toprule
\multirow{2}{*}{Dataset} & \multicolumn{2}{c}{Plain image} & \multicolumn{4}{c}{SeeEEG token-local strips} & \multirow{2}{*}{\begin{tabular}{@{}c@{}}Best\\EEG FM\end{tabular}} \\
\cmidrule(lr){2-3}\cmidrule(lr){4-7}
 & Traces & Matrix~\cite{li2026mtdp} & Stored & Clinical & Learned & Full & \\
\midrule
TUEV & 48.2 & 55.5 & 61.3 & 65.3 & \textbf{71.1} & 70.6 & 68.9 \\
TUAB & 75.9 & 77.7 & 77.6 & \textbf{81.5} & 79.6 & 81.2 & 80.7 \\
HMC & 62.7 & 65.1 & 71.0 & 71.0 & 70.9 & \textbf{72.0} & 68.5 \\
ADFTD & 40.9 & 47.7 & 46.3 & \textbf{59.5} & \textbf{59.5} & 54.1 & 51.9 \\
TUSL & 64.4 & 68.3 & 70.7 & 83.9 & 83.9 & \textbf{90.4} & 74.5 \\
SEED & 55.3 & 59.6 & 62.6 & 62.6 & \textbf{66.7} & 65.6 & 61.5 \\
BCIC2020-3 & 27.1 & 33.9 & 28.8 & 33.0 & 33.1 & \textbf{35.6} & 29.0 \\
PhysioMI & 29.7 & 38.9 & 36.3 & 36.3 & 46.5 & \textbf{47.2} & 45.1 \\
\midrule
Mean & 50.5 & 55.8 & 56.8 & 61.6 & 63.9 & \textbf{64.6} & 60.0 \\
\bottomrule
\end{tabular}
\end{adjustbox}
\end{table}

\Cref{tab:buildup} varies only the input image, with the frozen ViT, readout heads, and validation criteria fixed. Rendering
EEG as a standard page of traces, as in a conventional EEG viewer, yields a mean balanced accuracy of \buwavemean, and rendering it as a
channel-by-time signal matrix~\cite{li2026mtdp} yields \buvmtdpmean. Both fall short of the \bufmmean\ obtained by the best EEG
foundation model on each dataset. This gap is consistent with the design of MTDP~\cite{li2026mtdp}, which extracts frozen DINOv3 features from the
signal matrix but still distills them into an encoder pretrained on additional EEG data.
Token-local strips on the stored referential montage raise the mean to \bubothmean, and selecting the best clinical montage on validation raises it
to \buclinicalmean, which already lets the frozen ViT outperform every EEG foundation model on \bufirstclinical\ of the
eight datasets. The learned display raises the mean to \bulearnedmean, and the complete SeeEEG pipeline reaches \bufinalmean.
This score is \buplaingap\ points above the stronger plain image and \bufmgap\ points above the per-dataset best EEG foundation model,
and SeeEEG outperforms every foundation model on all \bufirstfinal\ datasets.

Different display components help different tasks. Because the stored reference is an artifact of acquisition and
preprocessing, the component shared by all channels accounts for \refcmadftd\ of the signal
variance on ADFTD. The common average reference removes this component and raises the dementia classification $\kappa$ from \dispadftdref\ to
\dispadftdcar. On slowing classification, the bipolar montage raises $\kappa$ from \disptuslref\ to \disptusltcp, whereas
the common average reference lowers it to \disptuslcar. This task depends on the local electrode contrasts captured by spatial derivations,
and the whitened display further raises $\kappa$ to \disptuslwhite. On motor imagery classification, the learned display adds
\bulearnedgainphysiomi\ points of balanced accuracy over the best clinical montage. The display also matters for EEG
foundation models: evaluating the montage selected on validation instead of the stored referential montage improves the dementia classification $\kappa$ of
\dispfmoneadftdname\ from \dispfmoneadftdref\ to \dispfmoneadftdsel\ and the slowing classification $\kappa$ of \dispfmonetuslname\ from
\dispfmonetuslref\ to \dispfmonetuslsel.

\subsection{Learned display}
\label{sec:learns}

\begin{figure*}[t]
\centering
\includegraphics[width=0.99\linewidth]{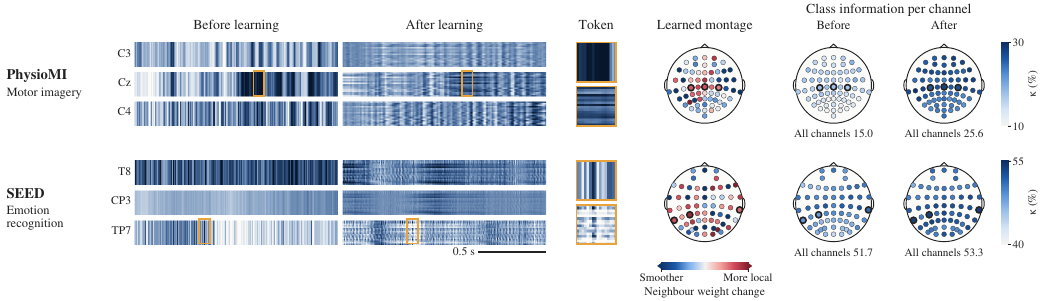}
\caption{\textbf{Learned displays.} For one validation window each from PhysioMI and SEED, the strips show the value view of three channels before and after optimization, and the token column enlarges the boxed token before (top) and after (bottom) optimization. The montage map shows how the neighboring weights of each channel change. The right maps report the validation $\kappa$ of a linear probe on the frozen tokens of each channel, with the strip channels circled and the probe on all channels reported below.}
\label{fig:learned}
\end{figure*}

\Cref{fig:learned} shows how the learned display changes the input image, with motor imagery in the top row. Before optimization,
the 16 pixel rows of a value strip replicate one waveform, so each token sees 16 copies of the same trace. After optimization,
each pixel row passes its channel through its own filter, and a single token represents the channel through 16 distinct filters.
C3, Cz, and C4 lie over the hand and foot regions of the motor cortex.
Starting from the stored referential montage, the display turns these electrodes into local derivations whose neighboring weights sum to $+0.90$, $+0.97$, and $+0.72$, respectively, close to
a surface Laplacian. It also averages most lateral and temporal electrodes with their neighbors and amplifies the overall contrast
by a factor of 2.2. The frozen tokens of the learned display carry more task-relevant information. On the validation windows, a linear
probe on the tokens of a single channel improves on all 64 channels. Its median $\kappa$ rises from 15.8 to 25.8 overall and
from 16.0 to 36.0 for left- versus right-hand imagery, and a probe on the mean token over all channels rises from 15.0
to 25.6. Emotion recognition, in the second row, shows the same pattern, with more class information in the frozen tokens
of 54 of the 60 channels. No electrodes, frequency bands, or filters were specified in advance,
and the display found these configurations through the frozen backbone from task labels alone.

\subsection{Image controls}
\label{sec:reads}

\begin{table}[t]
\centering
\footnotesize
\setlength{\tabcolsep}{2.6pt}
\caption{\textbf{Image controls.} Test score (\%, mean over three seeds) of SeeEEG on the original image and its change under each control. The head and view selected on the original image are retrained on each variant. The original image is the single display selected on validation, not the ensemble or the automatic sensitivity display.}
\label{tab:reads}
\begin{adjustbox}{max width=\linewidth}
\begin{tabular}{llcccccc}
\toprule
 & & & \multicolumn{2}{c}{patch alignment} & \multicolumn{2}{c}{token content} & pretraining \\
\cmidrule(lr){4-5}\cmidrule(lr){6-7}\cmidrule(lr){8-8}
dataset & metric & original & 16-px shift & 8-px shift & row cycle & time shuffle & random init \\
\midrule
TUEV & $\kappa$ & 82.5 & +1.6 & -3.7 & +2.3 & -1.2 & -15.9 \\
TUAB & AUROC & 88.1 & +0.0 & -0.1 & +0.3 & +0.6 & -1.0 \\
HMC & $\kappa$ & 66.3 & +0.7 & +0.5 & +0.8 & +0.0 & -7.4 \\
ADFTD & $\kappa$ & 40.2 & -0.9 & -1.8 & -6.9 & -6.4 & +0.0 \\
TUSL & $\kappa$ & 84.1 & -2.9 & -29.7 & -33.7 & -18.5 & -36.6 \\
SEED & $\kappa$ & 50.6 & +0.2 & +0.5 & +0.6 & -9.0 & -28.3 \\
BCIC2020-3 & $\kappa$ & 16.4 & +0.4 & -3.5 & -4.0 & -5.4 & -7.9 \\
PhysioMI & $\kappa$ & 28.6 & -1.5 & -3.0 & +1.3 & -10.4 & -24.6 \\
\midrule
mean $\Delta$ & & & -0.3 & -5.1 & -4.9 & -6.3 & -15.2 \\
\bottomrule
\end{tabular}
\end{adjustbox}
\end{table}

\begin{figure*}[t]
\centering
\includegraphics[width=0.99\linewidth]{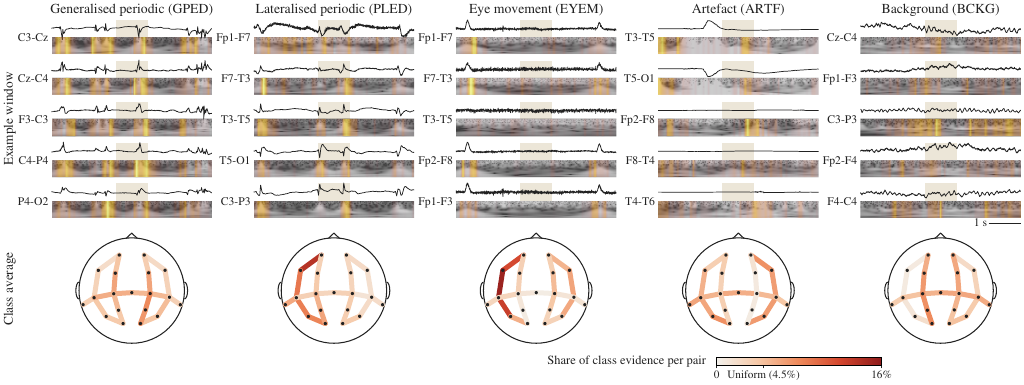}
\caption{\textbf{Evidence maps on TUEV.} The top row shows the most confidently recognized test window of each event type. The five bipolar pairs with the most evidence are drawn as traces above their scalogram strips, which are overlaid with the token-level evidence, and the shaded one-second interval marks the annotated event. The bottom row shows the share of evidence on each pair, averaged over the test windows of each class that the readout head recognizes.}
\label{fig:evidence}
\end{figure*}

\Cref{tab:reads} probes which visual properties the frozen ViT relies on by perturbing the rendered images of SeeEEG or randomly initializing the backbone,
retraining the readout head and view selected on validation for the original image under each condition.
A randomly initialized ViT-B with the same display and readout heads scores \rcrandgapmean\ points below DINOv3 on average
and up to \rcrandgapbest\ points below it on \rcrandgapbesttask\ classification. The ability to read rendered EEG comes from natural image pretraining. Shifting
the image vertically by a full token row keeps each strip aligned with the patch grid and changes the score by only \rcrollBGmean\
points on average. A half-row shift makes every token straddle two adjacent strips and degrades performance on \rcrolldown\
of the \rcn\ datasets, by up to \rcrollworst\ points on \rcrollworsttask\ classification. The frozen ViT relies on isolating each
channel within the tokens of its own strip.

Two further controls change the content of individual tokens. Cyclically shifting the 16 pixel rows of each strip by 8 rows keeps all pixel intensities and the
channel and time span of every token, and changes only the row order within each patch, which is the order of scalogram
frequencies and, in a learned display, of value-row filters. This permutation costs up to \rcfcycworst\ points on
\rcfcycworsttask\ classification and also lowers dementia classification and imagined speech recognition. On these
three tasks, whose discriminative evidence lies in the spectral profile, the frozen ViT reads the spectrum from its spatial layout within
each token, much as it reads local texture in an image patch. Shuffling the time columns keeps the marginal power at every frequency but
destroys temporal order. This perturbation degrades performance on \rcshufdown\ of the \rcn\ datasets and by \rcshufneg\ points on average,
so SeeEEG exploits temporal dynamics beyond the static spectrum.

\subsection{Evidence maps}
\label{sec:evidence}

\looseness=-1
The evidence maps in \cref{fig:evidence} show which spatial and temporal regions of the rendered image drive the prediction of the frozen ViT for each class. Because the linear
readout head acts on the CLS token concatenated with the mean patch token, the patch part of its logit decomposes exactly into one contribution per
token. In the most confidently recognized test window of each event type, positive evidence for periodic discharges coincides with the discharges themselves.
Averaged over the recognized test windows of each class, the evidence for lateralized discharges concentrates in one hemisphere, with a per-window asymmetry index $|L-R|/(L+R)$ of $0.45$ against $0.07$
for generalized discharges, and the frontal electrode pairs carry 24\% of the evidence for eye movement artifacts although they make up only 18\%
of the pairs. The readout head learned this physiologically grounded localization from window-level labels alone.

\section{Conclusion}
\label{sec:conclusion}

\looseness=1
A vision foundation model with no prior exposure to EEG can serve as a stronger frozen EEG encoder than specialized EEG foundation models, provided that each channel is rendered within a dedicated row of tokens and the display is learned through the frozen backbone. Consistent with the Platonic representation hypothesis~\cite{huh2024platonic}, this indicates that much of the structure that EEG foundation models aim to learn from scarce recordings is already encoded in models pretrained on natural images, and that adapting to a new EEG task primarily requires learning an appropriate display. EEG representation learning can benefit directly from future advances in vision foundation models, since a new backbone can be adapted to EEG by refitting only the lightweight display and readout head. This principle readily extends to other multichannel physiological signals that domain experts interpret through visual inspection, such as MEG and ECG.

\FloatBarrier
{
    \small
    \bibliographystyle{ieeenat_fullname}
    \bibliography{main}
}

\end{document}